\documentclass{article}
\usepackage{ijcai25}

\usepackage{times}
\usepackage{soul}
\usepackage{url}
\usepackage[hidelinks]{hyperref}
\usepackage[utf8]{inputenc}
\usepackage[small]{caption}
\usepackage{graphicx}
\usepackage{amsmath}
\usepackage{amsthm}
\usepackage{booktabs}
\usepackage{algorithm}
\usepackage{algorithmic}
\usepackage[switch]{lineno}
\usepackage{multirow}
\usepackage{amsfonts}
\usepackage{arydshln}
\usepackage{xcolor}
\usepackage{pifont}
\newcommand{\cmark}{\textcolor{green}{\ding{51}}} 
\newcommand{\xmark}{\textcolor{red}{\ding{55}}}   

\title{Critique Before Thinking:\\ Mitigating Hallucination through  Rationale-Augmented Instruction Tuning}

\author{Zexian Yang$^{1,2}$ \quad Dian Li$^{2}$\footnotemark[1] \quad Dayan Wu$^{1}$\footnotemark[1] \quad     Gang Liu$^{2}$ \quad Weiping Wang$^{1}$ \\
\affiliations $^{1}$Institute of Information Engineering, Chinese Academy of Sciences \\ $^{2}$Foundation Technology Center, Tencent PCG\\
}

\begin{document}

\maketitle
\renewcommand{\thefootnote}{\fnsymbol{footnote}} 
\footnotetext[1]{Co-corresponding authors.} 

\begin{abstract}
Despite significant advancements in multimodal reasoning tasks, existing Large Vision-Language Models (LVLMs) are prone to producing visually ungrounded responses when interpreting associated images. In contrast, when humans embark on learning new knowledge, they often rely on a set of fundamental pre-study principles: reviewing outlines to grasp core concepts, summarizing key points to guide their focus and enhance understanding. However, such preparatory actions are notably absent in the current instruction tuning processes. This paper presents Re-Critic, an easily scalable rationale-augmented framework designed to incorporate fundamental rules and chain-of-thought (CoT) as a bridge to enhance reasoning abilities. Specifically, Re-Critic develops a visual rationale synthesizer that scalably augments raw instructions with rationale explanation.  To probe more contextually grounded responses, Re-Critic employs an in-context self-critic mechanism to select response pairs for preference tuning. Experiments demonstrate that models fine-tuned with our rationale-augmented dataset yield gains that extend beyond hallucination-specific tasks to broader multimodal reasoning tasks.
\end{abstract}

\begin{figure}[t]
  \centering
  \includegraphics[width=0.95\linewidth]{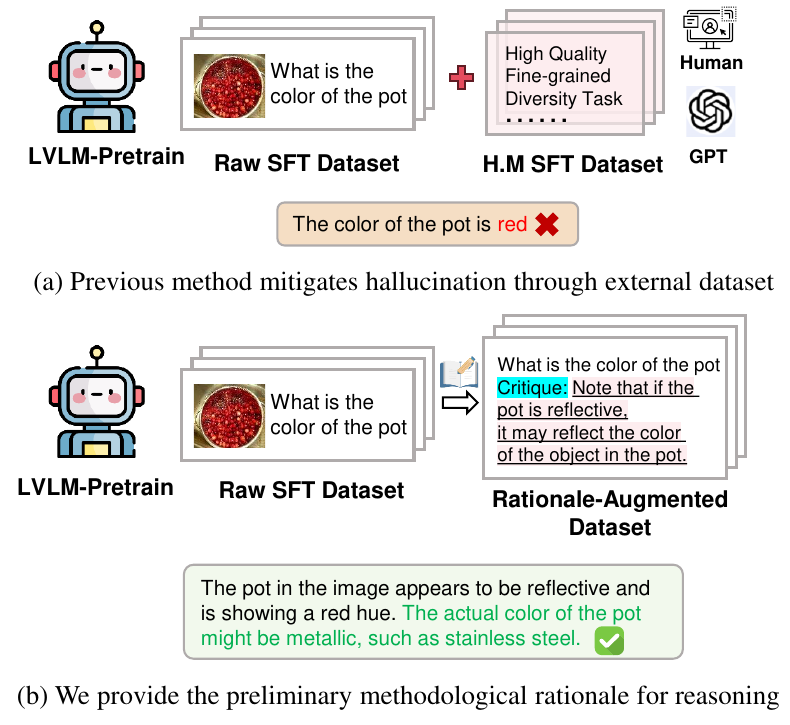}
  \caption{ Comparison of hallucination mitigation methods. In this example, our method provides the rationale explanation prior to reasoning, enabling more accurate interpretation of an object's color by considering its reflective properties. }
  \label{fig:1}
\end{figure}

\section{Introduction}
\label{submission}
The advent of Large Vision-Language Models (LVLMs), such as GPT-4V~\cite{chen2025sharegpt4v}, LLaVA~\cite{liu2024visual}, and InternVL~\cite{chen2024internvl} has achieved substantial progress in multimodal recognition tasks, including Image Captioning, Visual Question Answer (VQA), and Multimodal Conversation. Despite these advances, LVLMs tend to generate incorrect textual content that deviates from the factual content of the image, which is called \textit{hallucination} \cite{li2023evaluating}. This typically includes descriptions of non-existing visual contents and errors in descriptions. One cause of hallucinations in models stems from the ``Data Homogeneity"~\cite{liu2024survey}, which hinders the model's ability to understand visual information and execute instructions across diverse environments.

In response to this issue, many studies have been focused on collecting extra remedial training data. LRV-Instruction \cite{liu2023mitigating} proposes to diversify instructions to achieve more robust visual instruction tuning. M-HalDetect \cite{gunjal2024detecting} focuses on fine-grained annotations at
a sub-sentence level over detailed image descriptions. However, current instruction datasets mostly lack annotated explanations for the answers. Consequently, existing data-driven approaches predominantly rely on leveraging large-scale datasets to explore implicit correlations, neglecting the learning of logical reasoning and methodology. Moreover, the decision-making process of these models often lacks transparency and interpretability. We can vividly illustrate this point with a very intuitive example: imagine trying to learn mathematics without being taught the foundational theorems and principles. \textbf{If a teacher were to provide only a large number of problems and their solutions without explaining the underlying concepts, students would struggle to understand the subject truly.} They might be able to memorize some answers, but they would lack the critical thinking and problem-solving skills necessary to tackle new or complex problems independently. 

Based on this insight, we attribute the occurrence of hallucinations to insufficient context reasoning, specifically the lack of rationale guidance between the questions and answers. This deficiency causes the model to predominantly rely on the statistical patterns it learned during training, making it challenging for the model to generalize and apply knowledge effectively in new situations. For example, consider the thoughts one person might have when answering the question in Figure~\ref{fig:1}.  One first notices the material of the object and then recalls the knowledge regarding the behavior of reflective surfaces. By incorporating this fundamental principle as a prompt, we can encourage the model to apply more rigorous contextual reasoning, leading to more accurate and contextually grounded answers.

This paper presents Re-Critic, a rationale-augmented framework designed to incorporate fundamental rules and chain-of-thought (CoT) \cite{wei2022chain} as a bridge to enhance reasoning abilities. Specifically, Re-Critic encompasses two principal components: 1) the \textbf{V}isual \textbf{C}hain \textbf{I}nsertion \textbf{T}echnique (VCIT), which utilizes proprietary LLMs to generate rationales and 2) the self-critic mechanism that enables bootstrapped self-improvement by evaluating and refining the self-responses.  The VCIT component operates in two stages. Firstly, we employ a prompting strategy to guide proprietary LLMs in generating high-quality feedback. We utilize a prompt to tell proprietary LLMs to generate comprehensive rationales based on the provided question and reference answer. This process takes into consideration the contents of the associated image to ensure that the explanations are contextually relevant and accurate. The feedback can focus on key elements in the image that are critical for understanding the answer and provide a step-by-step reasoning process. Next, the LVLM undergoes training with rationale-augmented instructions, ensuring it adopts structured reasoning into its cognitive processes. Unlike conventional CoT reasoning methods that prompt the LVLM to produce rationales in response, we turn to incorporating the rationale context into the question itself, which mirrors the human recognition where foundational knowledge precedes answering.

Moreover, we further develop an effective and economical preference learning pipeline to enhance the alignment. This pipeline involves the self-critic mechanism, resulting in a high-quality dataset for Direct Preference Optimization (DPO) \cite{rafailov2024direct}. Compared to existing preference tuning methods, our solution offers two primary advantages: 1) Third-part API free: instead of depending on external AI models for feedback or reward models to assess response quality, it employs the model’s own judgments to evaluate the quality of responses. 2) Self-critic avoids distribution shift issues: by utilizing the model's own judgments to assess response quality, we mitigate the potential biases and variations that might arise from integrating feedback from external sources. 

We demonstrate the effectiveness of Re-Critic across a wide range of hallucinations and comprehensive benchmarks. Re-Critic achieves notable improvements on hallucination-specific benchmarks such as MMHal \cite{sun2023aligning}, HallusionBench \cite{guan2024hallusionbench}, and POPE \cite{li2023evaluating}, as well as on multimodal reasoning benchmarks including MME \cite{fu2023mme} and MathVista \cite{lu2023mathvista}. Remarkably, by fine-tuning the LLaVA-80K instruct dataset, we observe a performance enhancement of 31.8\% compared to baseline methods using an equivalent amount of data. This demonstrates that the model can compensate for data insufficiency through effective learning methodologies. We summarize the contribution of this paper as follows:

\begin{itemize}

\item Orthogonal to curating external high-quality datasets for hallucination mitigation, we underline the significance of methodological learning before reasoning. These insight highlight the shortcomings of existing  methods in efficiently bridging the visual rationale between instructions and responses.

\item  We propose a scalable framework called Re-Critic, specifically designed for mitigating hallucinations.  Re-Critic introduces a visual chain insertion technique, which effectively enhances visual reasoning capabilities through structured  rationale integration.

\item Our experiments show that equipping LVLMs with Re-Critic not only mitigates hallucinations but also effectively improves the performance across multiple benchmark evaluations.
\end{itemize}

\begin{figure*}[t]
  \centering
  \includegraphics[width=\linewidth]{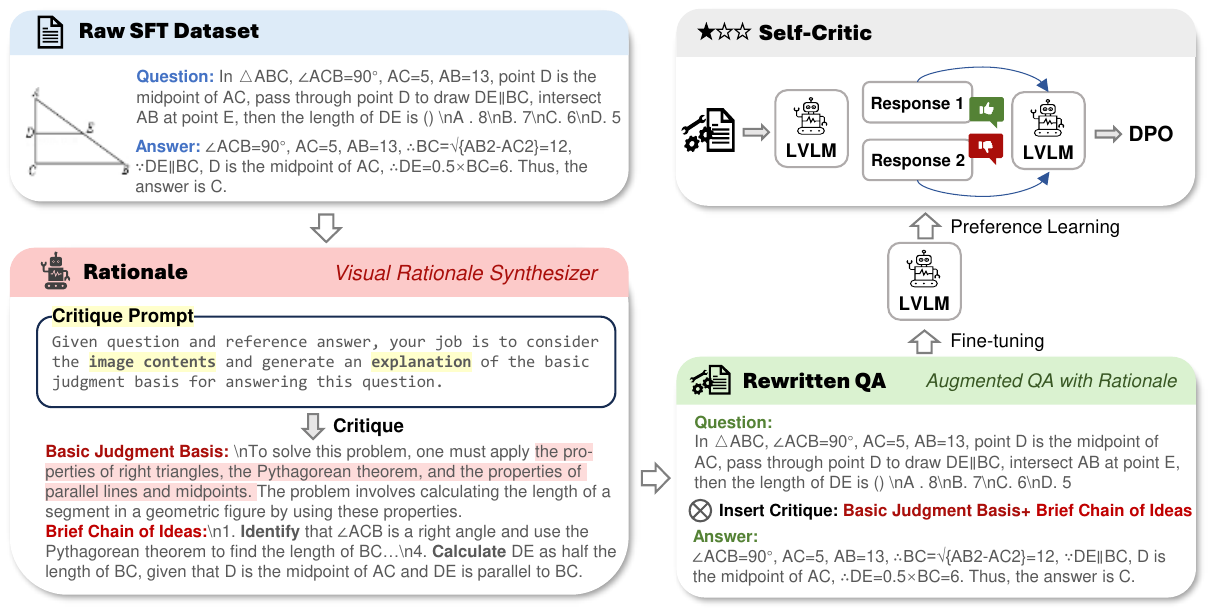}
  \caption{ \textbf{Overview of Re-Critic.}  The overall process is divided into three parts: 1) augmented standard QA with synthesized visual rationales, then 2) fine-tuning LVLM with rewritten QA, and finally 3) perform DPO through the self-critic preference learning. }
  \label{fig:2}
\end{figure*}
\section{Related Work}

\textbf{Hallucination in LVLMs.}
Hallucination remains a significant challenge in LVLMs, where models generate information that is factually incorrect or irrelevant to the given inputs. In multimodal reasoning, hallucinations can occur when models produce textual outputs not grounded in the visual data, leading to fabricated details in tasks such as image captioning or visual question answering (VQA). For hallucination evaluation, FaithScore \cite{jing2023faithscore} extracts fine-grained atomic facts from the generated answer and then conducts consistency verifications. POPE \cite{li2023evaluating}, a polling-based query method is proposed for a better evaluation of object hallucination. HallusionBench \cite{guan2024hallusionbench} recently explored the hallucination evaluation, by editing the original input image and forming different text-image pairs to diagnose failure types from language hallucination and visual illusion. For hallucination mitigation within LVLMs, previous works generally either collect more high-quality data manually or attach an extra correction model. Our method, however, shifts the attention to extended visual reasoning chains in advancing model reasoning capabilities.

\textbf{Reasoning with LLMs.} A large body of research \cite{qiao2022reasoning} has shown that the reasoning capabilities of LLMs can be enhanced through advanced inference-time techniques, such as prompting and aggregation. Several studies \cite{jiang2024forward}  have advocated for the use of backward reasoning to validate the chain of thought and improve mathematical reasoning. While effective, these methods are primarily applied at test time and rely heavily on the inherent capabilities of the LLMs. This dependency may limit the broader applicability and scalability of these approaches.

\textbf{Hallucination Reduction with Alignment.}
Several alignment techniques are utilized to better match LVLMs with human preferences. One commonly employed method is Reinforcement Learning from Human Feedback (RLHF), which iteratively improves the model's responses based on human feedback. LLaVA-RLHF \cite{sun2023aligning} harnesses the expertise of human evaluators to mark the more hallucinated output, thereby reducing the incidence of hallucinations.  Beyond RLHF, Direct Preference Optimization (DPO) has emerged as a promising technique to streamline the alignment process. RLHF-V \cite{yu2024rlhf} first employs DPO and develops a fine-grained correctional human feedback learning framework for behavior alignment. RLAIF-V \cite{yu2024rlaif} further refines DPO using high-quality responses from multiple LVLMs. However, these methods utilize the third-part feedback and consequently cause the distribution shift problem. Our work is primarily centered on further improving the alignment between the vision and text modalities through self-evaluation alignments.

\section{Method}
Re-Critic is a rationale-augmented framework that enables LVLMs to learn methodologies via instruction tuning, thereby facilitating advanced reasoning.  Figure 2 shows examples of how Re-Critic works. Given a multimodal question that includes both visual and textual components, Re-Critic augments each text from the original corpus with instruction-rationale pairs generated by a visual rationale synthesizer. These instruction-response pairs are used as inputs for tasks, and the LVLMs are subsequently fine-tuned using this enriched corpus.

\subsection{Visual Rationale Synthesizer}
To improve the reasoning ability via robust instruction tuning, our core idea is to encourage the model to apply visual contextual reasoning. To this end, we develop a visual rationale synthesizer to generate instruction-rationale pairs based on the content of the multimodal query. Considering the large amount of data required for supervised fine-tuning (SFT),  we start by collecting data related to reasoning from the original SFT datasets, rather than generating new data. This includes a wide range of tasks that often require strong visual and prior reasoning knowledge: (1) general visual conversation, detailed captioning, and reasoning (LLaVA-Instruction-150k). (2) various specific domains such as geometry problems (GeoQA+) and chart reasoning (ChartQA).

Then, we apply the visual chain insertion technique (VCIT) to multimodal queries from collected SFT datasets. As illustrated in Figure~\ref{fig:2}, based on each raw corpus, including both question and reference answer, we ask GPT4o to consider the image contents and exploit the gold answer to generate an explanation of the key judgment basis for answering this question. Unlike previous methods that include feedback as part of the answer, the generated rationale feedback will be inserted into the original question to construct an augmented input. During instruction tuning, we mix the collected instruction-rationale pairs with the rest SFT datasets, and the training sample is formulated as: 
\begin{equation}
 \{\text{(Image}~\mathbf{x}^I, \text{Question}~\mathbf{x}^T, \textbf{Rationale}~\mathbf{x}^R), \text{Answer}~\mathbf{y}^T \},
    \label{eq:1}
\end{equation}
where the \textbf{bold} part is specified for the augmented sample, and we adopt the auto-regressive training paradigm to fine-tune the model.  The learning objective is as followed:

\begin{equation}
    \mathcal{L}=-\sum_k{\log{P(y_{k+1}^T|y_k^T,(\mathbf{x}^I,\mathbf{x}^T,\mathbf{x}^R);\theta )}},
\end{equation}
where the $\theta$ denotes the learnable weights of the model.

\subsection{Self-Critic Prefernece Learning}
The goal of preference alignment is to ensure that the outputs of current LVLMs  better align with human preferences.  The feedback for preference learning is structured as comparison pairs, with each pair consisting of faithful responses,  $y_w$, and a less desirable response, $y_l$, both corresponding to the same input. During optimization, the model learns to identify and differentiate between $y_w$ and $y_l$, thereby understanding and internalizing the preferences. Previous approaches \cite{yu2024rlaif,zhao2023beyond} typically necessitate the use of external models to generate preference datasets. However, the substantial distribution shift between the external models and the LVLMs being optimized might render the generated datasets less effective. 

In response, we introduce a self-critic preference learning pipeline. The key innovation lies in the evaluation and refinement of self-generated responses without the need for an additional reward model. As shown in Figure~\ref{fig:2}, given the formatted input augmented with rationale $\mathbf{x}_a =(\mathbf{x}^I,\mathbf{x}^T,\mathbf{x}^R)$, we use the currently optimized model to generate two different response candidates for subsequent ranking and preference tuning. As to the response evaluation, we directly input the self-generated responses and the critic prompt into the currently optimized LVLM. The LVLM evaluates the responses, identifies the superior response as the positive one, and designates the other as the negative response.  
The subsequent challenge is to design an appropriate critic prompt, as the quality of the critic directly impacts the performance of the LVLM when optimized using the response pairs. The details of the critic metrics are outlined as follows:
\begin{itemize}
\item \textbf{Image Content Understanding}: Analogous to most hallucination detection methods, the critic  
should consider the accuracy of image content at various levels, including objects, attributes, and relationships
\item \textbf{Comprehensive Contextual Reasoning}:  Review both responses and compare the answers, prioritizing the response that demonstrates a more thorough understanding of the context. Pay special attention to results that are presented as a sequence of reasoning steps. It is important to verify the accuracy of each step, as some cases may have correct final answers but contain errors in the intermediate reasoning steps.

\end{itemize}
After obtaining the preference pairs through self-critic, we use these preference pairs to perform preference tuning on the current LVLM. While various frameworks exist for alignment training, we opt for Direct Preference Optimization (DPO) to achieve context-faithful alignment. Concretely, DPO enables the policy $\pi_\theta$ to be learned.  Given the input prompt $x$, the DPO dataset is construct as $\mathcal{D}=\{(\mathbf{x}_a, y_w,y_l) \}$. Then, we optimize the current model to minimize $\mathcal{L}_\text{DPO}(\pi_\theta;\pi_\text{ref})$, which can be formulated as follows:

\begin{equation}
\mathbb{E}_{(\mathbf{x}_a, y_w,y_l) \sim \mathcal{D}} [ -\log \sigma ( r(y_w \mid\mathbf{x}_a) - r(y_l \mid \mathbf{x}_a))],
\end{equation}
where $r(\cdot\mid x) =  \beta \log  \frac{\pi_\theta (\cdot\mid x)}{\pi_\text{ref} (\cdot\mid x)}  $ and $\beta$ represents a coefficient that  adjusts the influence of $\pi_\theta$ relative to  $\pi_\text{ref}$.

\begin{table}[t]
    \centering
    \resizebox{\linewidth}{!}{\begin{tabular}{lccccc}
    \toprule \hline
         & Ours &  LRV &  RLHF & HA-DPO &  POVID \\ \hline
      Construct Extra Instructions?   & \xmark  & \cmark &\cmark  &\cmark  & \xmark  \\
      Augmented Raw Instructions?   & \cmark  & \xmark &\xmark  &\xmark  & \xmark  \\
      External Reward Model? & \xmark & \xmark &  \cmark & \cmark & \cmark \\
      Num of Augmented Datasets  & 10k-30k & \xmark &\xmark & \xmark & \xmark \\
      Num of Curated Datasets  & \xmark & 400k & 120k & 28k & \xmark \\
   \hline \bottomrule
    \end{tabular}}
    \caption{A comparison of Re-Critic with datasets used by current methods.}
    \label{tab:1}
\end{table}
\section{Experiments}
\subsection{Experimental Setup}

Our Re-Critic is applicable to various LVLMs. The primary objective of the experiment is to demonstrate the effectiveness of our rationale-augmented framework using the same dataset across different LVLMs. We mainly chose LLaVA-v1.5-7B \cite{liu2024improved} and InternVL2-2B \cite{chen2024far} as the backbone models to validate the effectiveness, due to the availability of their code and models. For the LLaVA-v1.5 backbone, we use LLaVA-Instruct-665k as the raw SFT dataset and select a 10k subset for augmentations. Following the previous work \cite{huang2024mini}, the training datasets for InternVL2-2B include GeoQA+ \cite{cao2022augmented}, DocVQA \cite{mathew2021docvqa}, ChartQA \cite{masry2022chartqa}, DVQA \cite{kafle2018dvqa}, AI2D \cite{kembhavi2016diagram}, and LLaVA-150K (zh). We selectively chose 10\% of the data for augmentation, specifically targeting sub-tasks such as mathematics and science.

\noindent{\textbf{Benchmarks.}} We primarily conduct the experiments on four hallucination benchmarks: 

(1) \textbf{POPE} \cite{li2023evaluating} stands as a popular benchmark for assessing object hallucination using discriminative tasks. It employs binary question-answer pairs to prompt LVLMs to ascertain whether a specified object is present or absent in a given image;

(2) \textbf{MMHalBench} \cite{sun2023aligning}  evaluates hallucination across 12 different object-related topics, such as object attributes, the presence of adversarial objects, and spatial relations, among others. It utilizes GPT-4 to compare the model outputs with human responses for evaluation.

(3) \textbf{HallusionBench} \cite{guan2024hallusionbench} assessing visual illusion and knowledge hallucination with systematically structured discriminative tasks; 

(4) \textbf{Object HalBench} \cite{rohrbach2018object} is a popular benchmark utilized to evaluate the occurrence of common object hallucinations in detailed image descriptions. Following \cite{yu2024rlhf}, we employ 8 varied prompts to enhance the consistency of our assessment. We present both the hallucination rate at the response level (i.e., the proportion of responses containing hallucinations) and at the mentioned level (i.e., the proportion of objects that are hallucinated).

We also test the performance on seven general evaluation benchmarks, MME \cite{fu2023mme}, MathVista \cite{lu2023mathvista}, LLaVA-Bench (LLaVA-B) \cite{liu2024visual}, MMBench \cite{xu2023mmbench}, RealWorldQA(RWQA), GQA \cite{hudson2019gqa} and OCRBench \cite{liu2023hidden}.

\begin{table*}[t]
\centering
\resizebox{0.95\linewidth}{!}{\begin{tabular}{llllllll}
\toprule\hline
\multicolumn{1}{l}{\multirow{2}{*}{\textbf{Model}}}                              &\multicolumn{1}{c}{\multirow{2}{*}{\textbf{Parameter}}}                   & \multirow{2}{*}{\textbf{POPE}}   & \multicolumn{2}{c}{\textbf{MMHal-B}} &\multirow{2}{*}{\textbf{HallusionBench}} & \multicolumn{2}{c}{\textbf{Object HalBench}}                            \\ \cline{4-5} \cline{7-8} 
\multicolumn{1}{c}{}                                   & \multicolumn{1}{c}{}                      &                         & \multicolumn{1}{c}{Score}        & Hall. $\downarrow$        &      & Resp.$\downarrow$  & Ment.$\downarrow$ \\ \hline
\textit{Prior Multimodal LLMs}                         &                                            &                         &              &              &           &                                &                                \\ \hline
Gemini 1.5 pro \cite{team2024gemini}                                        & \multicolumn{1}{l|}{-}                     & \textbf{88.2}                   & -            &      -        & 45.6      & -                              & -        \\
GPT-4v \cite{achiam2023gpt}                                                & \multicolumn{1}{l|}{-}                      & 75.4                    & \textbf{3.49}         &    \textbf{0.28}          & \textbf{46.5}      & \textbf{13.6}                           & \textbf{7.3}      \\
Qwen-VL-Chat \cite{bai2023qwen}                                          & \multicolumn{1}{l|}{10B}                   & 74.9                    & 2.76         &   0.39           & 36.8      & 43.8                           & 20.0      \\
DeepSeek-VL \cite{lu2024deepseek}                                           & \multicolumn{1}{l|}{2B}                    & 85.9                    & -            &   -           & 27.6      & 16.7                              & 9.6        \\
MiniCPM-V-2 \cite{yao2024minicpm}                                           & \multicolumn{1}{l|}{2.8B}                  & 86.3                    & -            &   -          & 36.1      & 14.5                              & 7.8      \\ \hline
\multicolumn{8}{l}{\textit{Methods Tailored for Hallucination Mitigation}}                             \\ \hline
LRV-Instruction \cite{liu2023mitigating}                                         & \multicolumn{1}{l|}{7B}                     &   80.0                  &          -    &     -        & 23.7      &        44.3                        &  24.1       \\
LLaVA-v1.5 \cite{liu2024improved}                                                 & \multicolumn{1}{l|}{7B}                    & 85.9                    &  2.07         &  59.4          & 27.6         & 56.4                          & 27.9      \\
+ LLaVA-RLHF \cite{sun2023aligning}                                             & \multicolumn{1}{l|}{7B}                     &     $80.8_ {\color[HTML]{CB0000}(\downarrow 5.1)}$                   &    $2.05_{\color[HTML]{CB0000}(\downarrow 0.02)}$          &  $68.1_{\color[HTML]{CB0000}(\uparrow 8.7)}$            &      -      &   $40.1_{\color[HTML]{009901}(\downarrow 16.3)}$                    & $20.4_{\color[HTML]{009901}(\downarrow 7.5)}$           \\
+ HA-DPO \cite{zhao2023beyond}                                                 & \multicolumn{1}{l|}{7B}                    & $84.3_{\color[HTML]{CB0000}(\downarrow 1.6)}$                    & $1.98_{\color[HTML]{CB0000}(\downarrow 0.09)}$         &  $60.4_{\color[HTML]{CB0000}(\uparrow 1.0)}$            & -         & $\textbf{39.9}_{\color[HTML]{009901}(\downarrow 16.5)}$                           & $\textbf{19.9}_{\color[HTML]{009901}(\downarrow 8.0)}$      \\
+ POVID  \cite{zhou2024aligning}                                                & \multicolumn{1}{l|}{7B}                     &  $86.3_{\color[HTML]{009901}(\uparrow 0.4)}$                       &       $2.08_{\color[HTML]{009901}(\uparrow 0.01)}$       &   $56.2_{\color[HTML]{009901}(\downarrow 3.2)}$           &     $27.9_{\color[HTML]{009901}(\uparrow 0.3)}$      &          $48.1_{\color[HTML]{009901}(\downarrow 8.3)}$                      &  $24.4_{\color[HTML]{009901}(\downarrow 3.5)}$       \\
+ \textbf{Re-Critic (Ours) }                                           & \multicolumn{1}{l|}{7B}                      &    $\textbf{86.5}_{\color[HTML]{009901}(\uparrow 0.6)}$                     &     $\textbf{2.21}_{\color[HTML]{009901}(\uparrow 0.14)}$         &     $\textbf{56.0}_{\color[HTML]{009901}(\downarrow 3.4)}$         &    $\textbf{28.1}_{\color[HTML]{009901}(\uparrow 0.5)}$       &            $49.6_{\color[HTML]{009901}(\downarrow 6.8)}$                    &    $24.9_{\color[HTML]{009901}(\downarrow 3.0)}$       \\\hdashline
InternVL 2.0                                         & \multicolumn{1}{l|}{2B}                      &    85.2                     &   2.35           &    56.0          &    37.9       &                          13.8 & \textbf{7.1}         \\
+ \textbf{Re-Critic (Ours) }                                           & \multicolumn{1}{l|}{2B}                      &   $\textbf{88.2}_{\color[HTML]{009901}(\uparrow 3.0)}$                      &    $\textbf{2.35}_{\color[HTML]{C0C0C0}(\uparrow 0.00)}$           &    $\textbf{55.0}_{\color[HTML]{009901}(\downarrow 1.0)}$          &   $\textbf{39.1}_{\color[HTML]{009901}(\uparrow 1.2)}$        &                               $\textbf{12.9}_{\color[HTML]{009901}(\downarrow 0.9)}$ & $7.2_{\color[HTML]{CB0000}(\uparrow 0.1)}$       \\ \hline \bottomrule
\end{tabular}}
\caption{Performance comparison between Re-Critic and other baselines on on four hallucination benchmarks. Obj HalBench (Res./Men.) : Object HalBench with response/mention-level hallucination rates. The best performance for each category and task is in \textbf{bold}. The performance improvements compared to the vanilla model is denoted by green color.}
\label{tab:2}
\end{table*}

\begin{table*}[t]
\centering
\resizebox{\linewidth}{!}{\begin{tabular}{lllllllll}
\toprule \hline
\multirow{2}{*}{\textbf{Model}} & \multirow{2}{*}{\textbf{Parameter}} & \multirow{2}{*}{\textbf{MME}} & \multirow{2}{*}{\textbf{MathVista}} & \multirow{2}{*}{\textbf{LLaVA-B}} & \multirow{2}{*}{\textbf{MMBench}} & \multirow{2}{*}{\textbf{RWQA}} & \multirow{2}{*}{\textbf{GQA}} & \multirow{2}{*}{\textbf{OCRBench}} \\
                                &                                     &                               &                                     &                                   &                                   &                                &                               &                                    \\ \hline
\multicolumn{9}{l}{\textit{Prior Multimodal LLMs}}                                                                                                                                                                                                                                                                        \\ \hline
Gemini 1.5 pro                  & \multicolumn{1}{l|}{-}              & \textbf{2110.6 }                       & \textbf{57.7}                                & \textbf{95.3}                              & 73.9                              & \textbf{64.1 }                          & -                             & \textbf{754}                                \\
GPT-4v                          & \multicolumn{1}{l|}{-}              & 1771.5                        & 48.7                                & 93.1                              & \textbf{77.0}                              & 56.5                           & -                             & 516                                \\
Qwen-VL-Chat                    & \multicolumn{1}{l|}{10B}            & 1860.0                        & 34.9                                & 67.7                              & 61.8                              & 49.3                           & 60.7                          & 488                                \\
DeepSeek-VL                     & \multicolumn{1}{l|}{2B}             & 1531.6                        & 29.4                                & 51.1                              & 66.4                              & 49.7                           & 53.8                          & 413                                \\
MiniCPM-V-2                     & \multicolumn{1}{l|}{2.8B}           & 1808.6                        & 38.7                                & 69.2                              & 69.1                              & 55.8                           & \textbf{73.2}                          & 605                                \\ \hline
\multicolumn{9}{l}{\textit{Methods Tailored for Hallucination Mitigation}}                                                                                                                                                                                                                                                \\ \hline
LLaVA-v1.5                      & \multicolumn{1}{l|}{7B}             & \textbf{1858.9}                        & 25.4                                & 64.4                              & 64.3                              & 54.8                           & 62.0                          & 318                                \\
+ LLaVA-RLHF                    & \multicolumn{1}{l|}{7B}               & $1825.6_{\color[HTML]{CB0000}(\downarrow 33.3)}$                        & $23.5_ {\color[HTML]{CB0000}(\downarrow 1.9)}$                            & $61.5_{\color[HTML]{CB0000}(\downarrow 2.9)}$                             & $63.4_{\color[HTML]{CB0000}(\downarrow 0.9)}$                              & -                              & $61.3_{\color[HTML]{CB0000}(\downarrow 0.7)}$                          & $280_{\color[HTML]{CB0000}(\downarrow 38)}$                                \\
+ HA-DPO                        & \multicolumn{1}{l|}{7B}             & $1816.5_{\color[HTML]{CB0000}(\downarrow 42.4)}$                        & $25.8_{\color[HTML]{009901}(\uparrow 0.4)}$                              & $67.2_{\color[HTML]{009901}(\uparrow 2.8)}$                              & $64.9_{\color[HTML]{009901}(\uparrow 0.6)}$                              & -                              & $61.1_{\color[HTML]{CB0000}(\downarrow 0.9)}$                          & $311_{\color[HTML]{CB0000}(\downarrow 7)}$                                \\
+ POVID                         & \multicolumn{1}{l|}{7B}               & $1778.1_{\color[HTML]{CB0000}(\downarrow 80.8)}$                        & $24.4_{\color[HTML]{CB0000}(\downarrow 1.0)}$                                & $62.2_{\color[HTML]{CB0000}(\downarrow 2.2)}$                              & $64.9_{\color[HTML]{009901}(\uparrow 0.6)}$                              & -                              & $61.7_{\color[HTML]{CB0000}(\downarrow 0.3)}$                          & $312_{\color[HTML]{CB0000}(\downarrow 6)}$                                  \\ 
+ \textbf{Re-Critic (Ours) }                                           & \multicolumn{1}{l|}{7B}                      &     $1855.5_{\color[HTML]{CB0000}(\downarrow 3.4)}$                    &    $\textbf{27.2}_{\color[HTML]{009901}(\uparrow 1.8)}$          &       $\textbf{70.6}_{\color[HTML]{009901}(\uparrow 6.2)}$       &       $\textbf{67.4}_{\color[HTML]{009901}(\uparrow 3.1)}$    &   $\textbf{55.4}_{\color[HTML]{009901}(\uparrow 0.6)}$                              &    $\textbf{62.6}_{\color[HTML]{009901}(\uparrow 0.6)}$ & $\textbf{320}_{\color[HTML]{009901}(\uparrow 2)}$      \\ \hdashline
InternVL 2.0                                         & \multicolumn{1}{l|}{2B}                      &  1874.1
                        &     42.9         &    60.0        &      \textbf{73.1}     &         57.0                       &  \textbf{58.4} & 784        \\
+ \textbf{Re-Critic (Ours) }                                           & \multicolumn{1}{l|}{2B}                      &   $\textbf{1895.0}_{\color[HTML]{009901}(\uparrow 20.9)}$                      &      $\textbf{48.2}_{\color[HTML]{009901}(\uparrow 5.3)}$        &  $\textbf{66.2}_{\color[HTML]{009901}(\uparrow 6.2)}$     &  $73.0_{\color[HTML]{CB0000}(\downarrow 0.1)}$      &    $\textbf{58.0}_{\color[HTML]{009901}(\uparrow 1.0)}$       &      $58.0_{\color[HTML]{CB0000}(\downarrow 0.4)}$       &    $\textbf{803}_{\color[HTML]{009901}(\uparrow 19)}$       \\\hline \bottomrule
    \end{tabular}}
\caption{Experimental results on seven general multimodal benchmarks. RWQA: RealWorldQA, LLaVA-B: LLaVA-Bench (In-the-Wild). The best performance for each category and task is in \textbf{bold}. The performance improvements compared to the vanilla model is denoted by green color.} 
\label{tab:3}
\end{table*}

\noindent{\textbf{Baselines.}} We mainly compare Re-Critic with baseline models LLaVA-v1.5-7B and InternVL2-2B, as well as the state-of-the-art methods tailored for reducing hallucination, i.e., instruction tuning method LRV-Instruction \cite{liu2023mitigating} and preference optimization methods LLaVA-RLHF \cite{sun2023aligning}, HA-DPO \cite{zhao2023beyond} and POVID \cite{zhou2024aligning}. A detailed comparison of Re-Critic with the datasets used by current methods is shown in Table \ref{tab:1}.

\begin{figure*}[t]
  \centering
  \includegraphics[width=\linewidth]{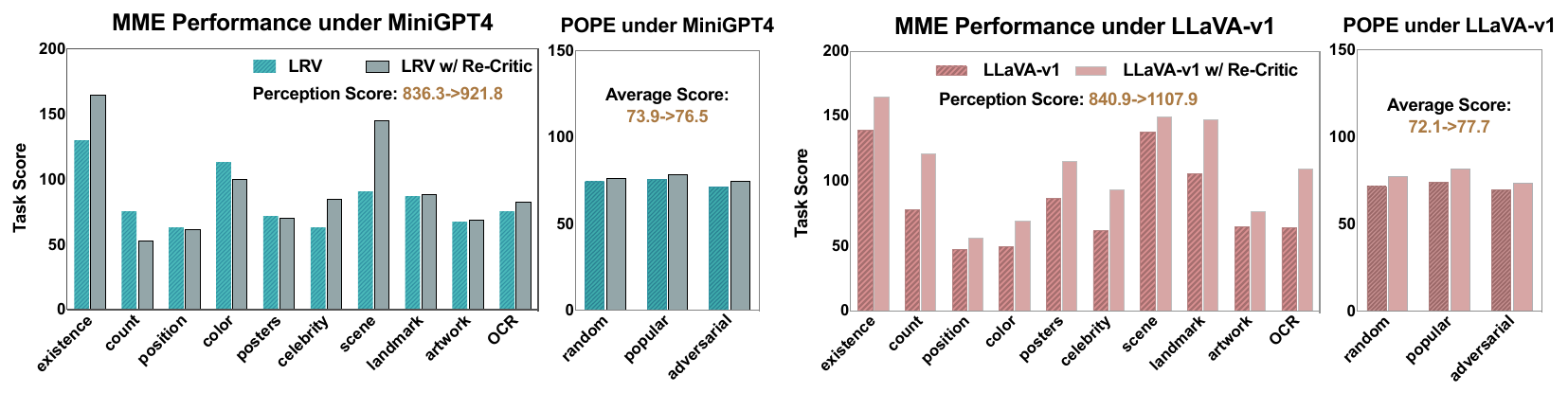}
  \caption{ Ablation study on the effectiveness of using the alternative base model MiniGPT4 (left) and smaller scale of training data (right). }
  \label{fig:3}
\end{figure*}

\begin{table*}[t]
\centering
\begin{tabular}{ccccccccl}
\toprule\hline
                       &                                            & \multicolumn{2}{c}{\textbf{MMHal-bench}}    &                        & \multicolumn{2}{c}{\textbf{Object HalBench}}                                         &                                  & \multicolumn{1}{c}{}                          \\ \cline{3-4} \cline{6-7}
\multirow{-2}{*}{\textbf{VCIT}} & \multirow{-2}{*}{\textbf{Self-Critic}}              & Score         & Hall. $\downarrow$ & \multirow{-2}{*}{\textbf{POPE}} & Resp. $\downarrow$                   & Ment. $\downarrow$                   & \multirow{-2}{*}{\textbf{HallusionBench}} & \multicolumn{1}{c}{\multirow{-2}{*}{\textbf{LLaVA-B}}} \\ \hline
\xmark  & \multicolumn{1}{c|}{\xmark} & 2.07          & 0.58               & 85.9                   & 56.4                                 & 27.9                                 & 27.6                             & 64.4                                          \\
\xmark  & \multicolumn{1}{c|}{\cmark} & 2.08          & 0.58               & 85.8                   & 56.7                                 & 26.7                                 & 28.1                             & \textbf{71.1}                                 \\
\cmark  & \multicolumn{1}{c|}{\xmark} & 2.18          & 0.57               & 86.5                   & {\color[HTML]{333333} 53.5}          & {\color[HTML]{333333} 25.0}          & 27.7                             & 70.2                                          \\
\cmark  & \multicolumn{1}{c|}{\cmark} & \textbf{2.21} & \textbf{0.56}      & \textbf{86.5}          & {\color[HTML]{333333} \textbf{49.6}} & {\color[HTML]{333333} \textbf{24.9}} & \textbf{28.1}                    & 70.6                                          \\ \hline \bottomrule
\end{tabular}
\caption{Ablation study on the effectiveness of each component of Re-Critic on LLaVA-v1.5 baseline. The best results in each column are highlighted in bold.}
\label{tab:4}
\end{table*}

\subsection{Main Results}
We benchmarked Re-Critic against the baseline methods, and the main results are summarized in Table \ref{tab:2} and Table \ref{tab:3}. Table \ref{tab:2} presents the results on four widely used hallucination benchmarks, while Table \ref{tab:3} provides the results on seven comprehensive benchmarks. We observe that our proposed Re-Critic not only excels on hallucination benchmarks but also demonstrates improved multimodal reasoning abilities on comprehensive benchmarks.

\textbf{Improvement with Re-Critic.} Proprietary multimodal models, such as GPT-4v and Gemini 1.5 Pro, achieve competitive results on both hallucination and general multimodal benchmarks. On the other hand, pioneering open-source models like LLaVA-v1.5 and InternVL 2.0 set a strong baseline in general vision-language tasks, but their performance on hallucination evaluation remains less than ideal.  Building upon these groundworks, Re-Critic consistently enhances the performance of base models across two types of benchmarks. Specifically, it achieves an average improvement of \textbf{6.2\%} for LLaVA-v1.5 and \textbf{2.7\% }for InternVL 2.0 in hallucination benchmarks. Additionally, it provides an average improvement of \textbf{2.8\%} for LLaVA-v1.5 and \textbf{3.7\%} for InternVL 2.0 on general benchmarks. In particular, we observe large gains in reasoning tasks. For instance, InternVL 2.0 with Re-Critic achieves an accuracy of 48.2\% on the mathematical benchmark MathVista. Furthermore, we observe improvements ranging from \textbf{1.7\%} to \textbf{10.3\%} on real-world benchmarks, indicating that the method enhances reasoning in complex scenarios.

\textbf{Compared with hallucination reduction methods.} 
We compare Re-Critic with current preference optimization baselines (i.e., LLaVA-RLHF, HA-DPO, POVID) on LLaVA-v1.5-7B. 
As shown, all the employed methods have yielded substantial enhancements on Object HalBench. However, the Re-Critic  has emerged as a superior performer, exhibiting significant improvements in performance across the remaining three datasets. 
Notably, both LLaVA-RLHF and HA-DPO underperform compared to Re-Critic on POPE and MMHal-Bench benchmarks and even compromise the general vision-language capabilities, resulting in performance that is not only inferior to Re-Critic but also falls short of the base model. A key distinction between these methods and Re-Critic lies in the integration of the visual chain-of-thought within the instruction datasets. Our results further underscore the significance of methodological learning before reasoning.  

\subsection{Ablation Study}

\textbf{Contributions from Different Components.} To assess the contribution of various components, we conduct ablation experiments by selecting different configurations. Table \ref{tab:4} summarizes the results. The results show that the self-critic preference learning demonstrates a impressive impact on fine-grained VQA tasks as indicated by the higher scores on the HallusionBench and LLaVA-Bench. This suggests its effectiveness in aligning LVLMs with human values to generate more accurate descriptions, without the need for third-party API intervention. Similarly, the introduction of the VCIT component shows notable improvements in reasoning and object hallucination tasks. Finally, when both VCIT and self-critic components are integrated, the model achieves the best performance on the POPE benchmark, as well as maintaining strong results in the MMHal-bench and Object HallusionBench.

\textbf{More LVLM Architectures.} To further validate the robustness of our approach, we also applied Re-Critic to other LVLM architectures. Specifically, we use the earlier and more rudimentary model, MiniGPT-4 \cite{zhu2023minigpt}, and select LRV-Instruction as the training data, with 10\% of the data used for rationale augmentation. We analyze performance across the comprehensive MME benchmark, which encompasses 10 subtasks designed to assess perception. Additionally, we evaluate hallucination using the POPE benchmarks based on three different metrics.  As illustrated in the left part of Figure \ref{fig:3}, Re-Critic significantly boosts performance across both benchmarks, leading to an average improvement of \textbf{10.2\%} in MME perception and \textbf{3.5\%} in POPE. Notably, LRV increases the proportion of negative instructions, and our approach further achieves larger improvements in the subtask ``existence" on MME. The results on MiniGPT-4 demonstrate that Re-Critic is general and effective for different LVLM architectures, regardless of the foundational capabilities of the base model.

\textbf{Impact of Using Smaller Scale Data.} While Re-Critic is effective with large-scale training data, we are also interested in how models trained on a smaller scale of data perform in general-purpose multimodal tasks. To this end, we employ LLaVA-Instruct-80K \cite{liu2024visual} as the visual instruction tuning data. Additionally, the rationale-augmented subset is constructed by randomly sampling 10K data from conversation and complex reasoning tasks. The results in the right part of Figure \ref{fig:3} show that 
using a smaller dataset generally leads to poorer performance. However, our method demonstrates a more significant performance improvement even with limited data. Specifically, our approach outperforms in each subtask on the MME benchmark, achieving an average performance gain of \textbf{31.8\%}. Besides, our approach achieves a substantial \textbf{7.8\% }gain on POPE. This suggests that Re-Critic can enhance the model's ability to learn effectively from limited data, highlighting the importance of learning methodologies prior to reasoning.

\begin{figure}[t]
  \centering
  \includegraphics[width=0.85\linewidth]{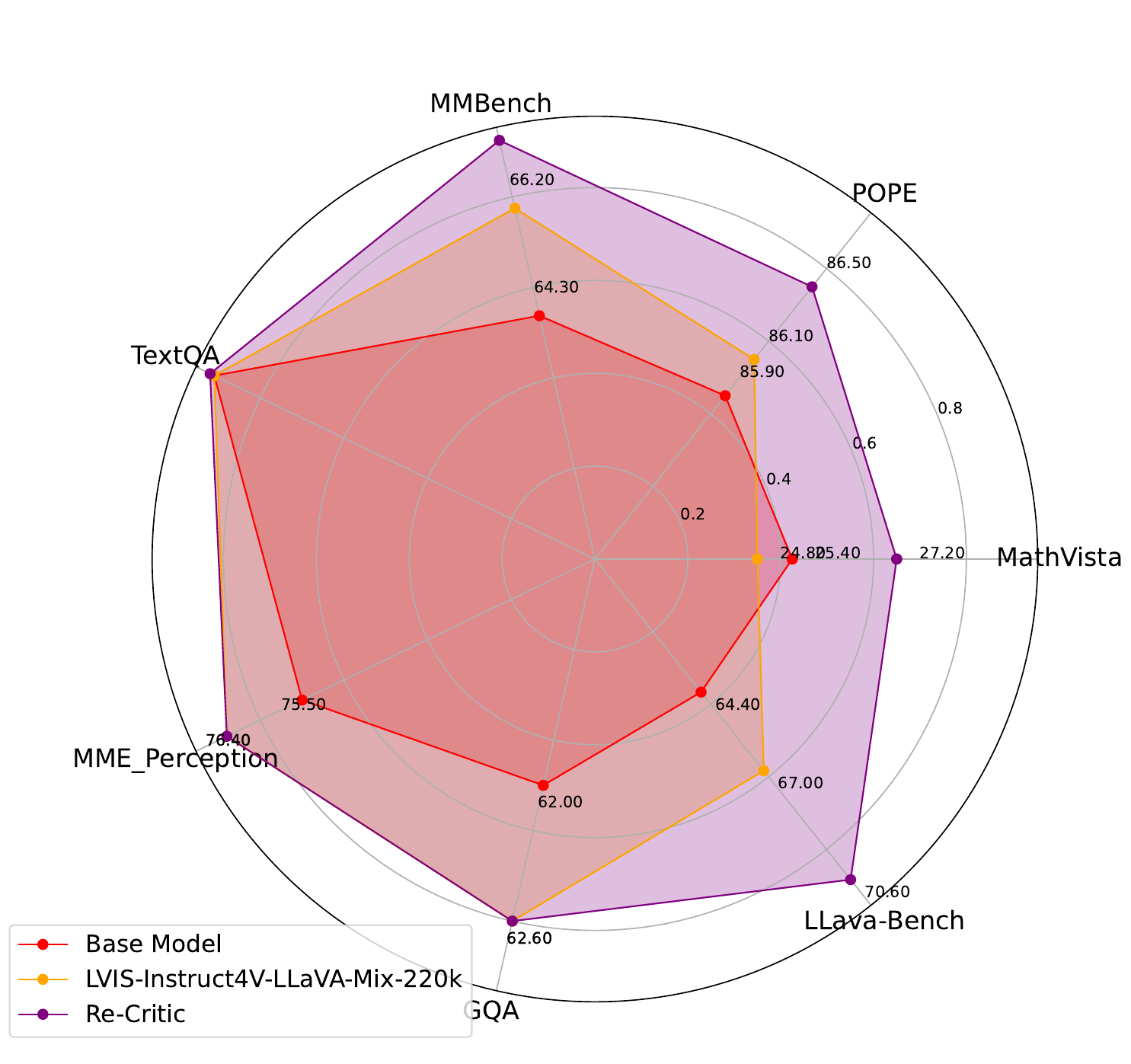}
  \caption{ Performance comparison of different models on six general tasks and one hallucination task.}
  \label{fig:4}
\end{figure}

\begin{table}[t]
\resizebox{\linewidth}{!}{\begin{tabular}{lcccc}
\toprule\hline
\textbf{Method }           & \textbf{MME}           & \textbf{RWQA}          & \textbf{MathVista}     & \textbf{OCRBench}     \\ \hline
Base Model        & 1874.1        & 57.0          & 42.9          & 802          \\
+ Random S. & 1882.8        & \textbf{58.6} & 47.4          & \textbf{803} \\
+ Hard S.   & \textbf{1895.0} & 58.0 & \textbf{48.2} & 802          \\ \hline\bottomrule
\end{tabular}}
\caption{Ablation study on the effectiveness of varying sampling strategies. The best results in each column are highlighted in bold.}
\label{tab:5}
\end{table}

\subsection{Further Analysis}

\textbf{Varying Sampling Strategies.} To investigate the impact of different sample enhancement strategies, Table \ref{tab:5} illustrates a comparison between two distinct mechanisms: random sampling and hard sample sampling.  For a fair comparison, we sampled 3.5k instructions from the mathematical dataset GeoQA+ using both strategies. In our approach to hard sample selection, inspired by active learning \cite{diao2023active}, we aim to identify and choose representative samples that have the potential to enhance the model's performance. The detailed methodology is designed as follows:  Initially, we utilized BERT-based sentence embeddings to cluster the questions. Next, we selected the top-k samples from each cluster to serve as representative examples. Finally, we conducted difficulty-based sorting to prioritize the types of questions where the model demonstrated lower accuracy. The performance of the random sampling strategy exhibits some improvements compared to the original base model. In contrast, the hard sample sampling strategy yields impressive improvements across all datasets, with particularly impressive gains observed in the MathVista dataset. These findings indicate that the hard sample sampling strategy may be more effective in enhancing contextual knowledge.

\textbf{Comparative Analysis of Data Augmentation and External Data Curation.} Figure \ref{fig:4} presents a comparison between our method and LVIS-INSTRUCT4V \cite{wang2023see}. Both methods adopt the same backbone model as LLaVA-1.5. LVIS, in particular, integrates the LLaVA-Instruct-665k dataset with their additional LVIS-INSTRUCT4V-220k dataset for visual instruction tuning. As depicted in the figure, our approach achieves comparable results to LVIS on TextQA, MME, and GQA. In addition, our method shows significant improvements on more challenging reasoning benchmarks, e.g., LLaVA-Bench and MathVista. This indicates that by employing effective data augmentation strategies, we can achieve performance on par with large-scale datasets without significantly increasing the data volume.

\begin{figure}[t]
  \centering
  \includegraphics[width=0.9\linewidth]{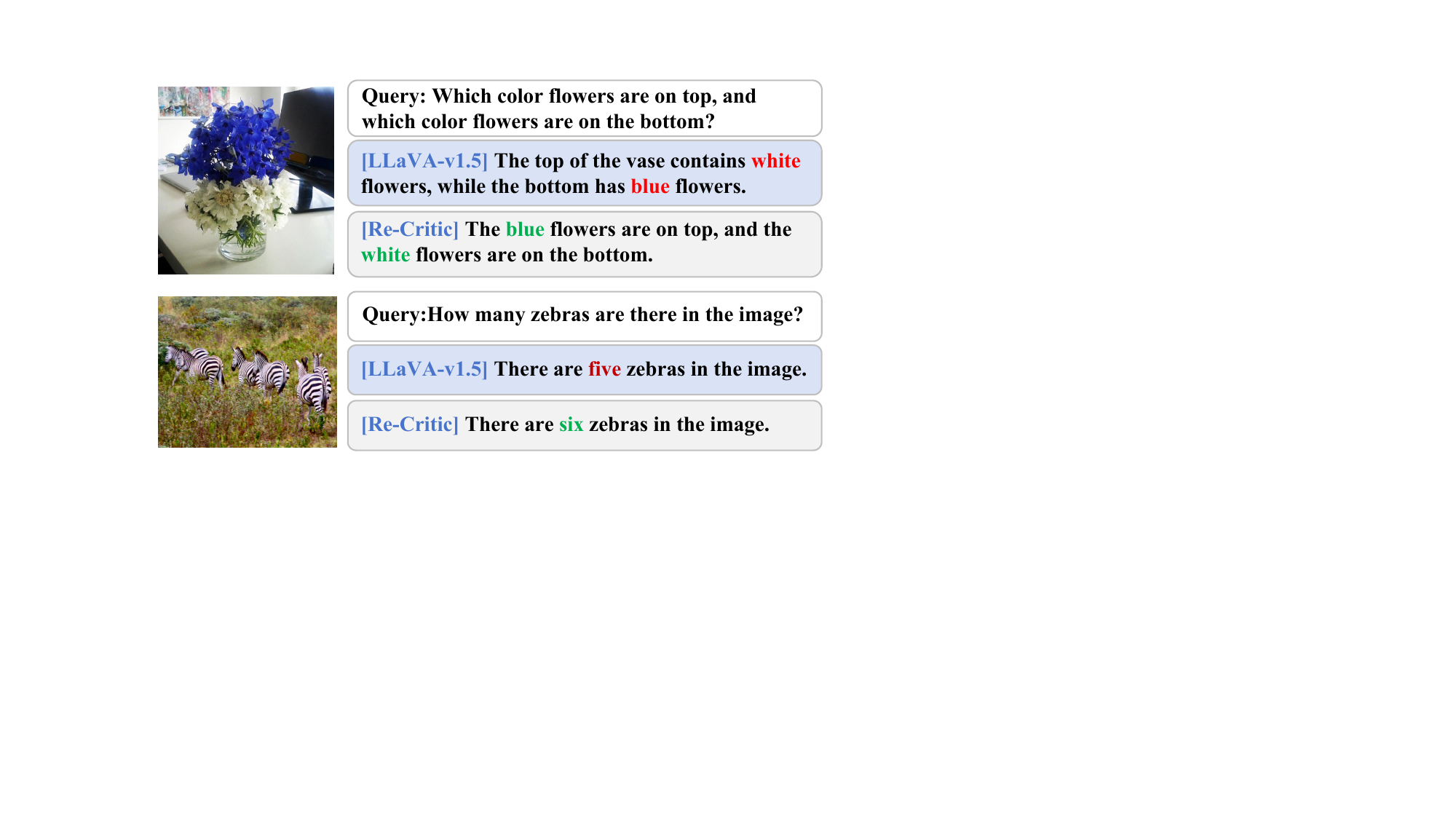}
  \caption{A case study of different predictions with LLaVA 1.5 as the backbone model.}
  \label{fig:5}
\end{figure}

\textbf{Case Study.} Figure \ref{fig:5} compares the different outputs of Re-Critic and the baseline on two examples from MMHal-Bench. The Re-Critic method effectively corrects the inaccuracies of the LLaVA-v1.5 baseline model, as demonstrated in two distinct examples. In the first instance, Re-Critic rectifies the misidentification of flower colors in a vase arrangement, confirming that the blue flowers are positioned at the top and the white flowers at the bottom. In the second example, Re-Critic accurately counts the number of zebras in an image, identifying six zebras instead of the five that LLaVA-v1.5 reported.

\section{Conclusion}

Existing instructions for training large multimodal models typically involve multi-modal inputs and output answers, leading to insufficient context reasoning in the models.  This paper represents an initial exploration into enabling large vision-language models (LVLMs) to possess basic knowledge reserves when undertaking reasoning tasks.  We introduce Re-Critic, an innovative rationale-augmented framework designed to enhance contextually grounded understanding in multimodal reasoning tasks. We demonstrate that incorporating visual rationales into instruction data is an effective way to teach the model the fundamental judgment criteria necessary to solve the problem.


\bibliographystyle{named}
\bibliography{ijcai25}

\end{document}